\documentstyle[11pt,latex-acl]{article}
\input{ijcaifig}
\author{Carolyn Penstein Ros\'{e} \\
Computational Linguistics Program \\ Carnegie Mellon University
 \\{\tt cprose@lcl.cmu.edu}\And
Alex Waibel \\
Dept.~of Computer
Science \\  Carnegie Mellon University \\
{\tt ahw@cs.cmu.edu}}

\title{Recovering From Parser Failures: A Hybrid
Statistical/Symbolic Approach}

\begin{document}
\maketitle
\bibliographystyle{acl}

\begin{abstract}

We describe an implementation of a hybrid statistical/symbolic
approach to repairing parser failures in
a speech-to-speech translation system.
\footnote{The research described in this paper was sponsored by
the Department of the Navy, Office of Naval Research, grant
\#N00014-93-1-0806.  The ideas described in this paper do not
necessarily reflect the position or the policy of the
government, and no official endorsement should be inferred.}
\footnote{This paper was originally published in the proceedings of
the Balancing Act workshop at the 32nd Annual Meeting of the ACL, Las Cruces,
New Mexico, 1994.}
We describe a
module which takes as input a fragmented parse and returns a
repaired
meaning representation.  It negotiates with the
speaker about what the complete meaning of the utterance is by
generating
hypotheses about how to fit the fragments of the partial parse
together into a coherent meaning representation.  By drawing upon
both
statistical and symbolic information, it constrains its repair
hypotheses to those which are both likely and meaningful.
Because it updates
its statistical model during use, it improves its performance
over time.

\end{abstract}

\section{Introduction}

Natural language processing of spontaneous speech is particularly
difficult because it contains false starts, out of vocabulary
words, and
ungrammatical constructions.  Because of this, it is unreasonable
to hope to be able to write a grammar which will cover all of the
phenomena
which a parser is likely to encounter in a practical speech
translation
system.  In this paper we describe an implementation of a hybrid
statistical/symbolic approach to recovering from parser failures
in the
context of a speech-to-speech translation system of significant scope
(vocabulary size of 996, word recognition
accuracy 60 \%, grammar size on the order of
2000 rules).
The domain
which
the current system focuses on is the scheduling domain where two
speakers
attempt to set up a meeting over the phone.

Because this is
an interlingua-based translation system, the goal of the analysis
stage of the translation process is to map the utterance in the
source language onto a feature-structure representation called an
interlingua
which represents meaning in a language-independent way.  (This approach
extends to other feature-structure based meaning representations as well.)
If the parser cannot
derive a complete analysis for an utterance, it derives a partial
parse by
skipping over portions of the utterance in order to find a subset
which can
parse.  It also returns an analysis for the skipped portions
which can be
used to rebuild the meaning of the input utterance.  The goal of
our repair
module is to interactively reconstruct the meaning of the full
utterance by
generating predictions about the way the fragments can fit
together and
checking them with the user.   In this way it negotiates with the
user in
order to recover the meaning of the user's utterance.

The repair module described in this paper uses both
symbolic and statistical information in order to reconstruct the
speaker's
meaning from the partial analysis which the parser produces.   It
generates
predictions based on constraints from a specification of the
interlingua
representation and from mutual information statistics extracted
from a corpus
of naturally occurring scheduling dialogues.  Mutual information is
intuitively
a measure of how strongly associated two concepts are.

Although the syntactic structure of the input utterance certainly plays an
important role in determining the meaning of an utterance, it is
possible
with the use of the interlingua specification to reason about the
meaning of
an utterance when only partial structural information is
available.  This can
be accomplished by fitting the partial features structures
together against
the mold of the interlingua specification.  During the parsing
process,
two structural representations are generated, one which is a
tree-like
structure generated from the structure of the context-free
portion of the
parsing grammar rules, and one which is a feature-structure
generated
from the unification portion of the parsing grammar rules.  There
is a
many-to-one mapping between tree-structures and
feature-structures.  Both
of these structures are important in the repair process.

The repair process is analogous in some ways to fitting pieces
of a
puzzle into a mold which contains receptacles for particular
shapes.  The
interlingua specification is like the mold with receptacles of
different
shapes, making it possible to compute all of the ways partial
analyses can
fit together in order to create a structure which is valid for
that
interlingua.  But the number of ways it is possible to do this
are so
numerous that the brute force method is computationally
intractable.  Mutual
information statistics are used to guide the search.  These
mutual information
statistics encode regularities in the types of fillers which tend
to occur
in particular slots and which feature-structures associated with
particular
non-terminal symbols in the parsing grammar tend to be used in a
particular
way in the interlingua representation.  By drawing upon both
statistical and
symbolic sources of information, the repair module can constrain
its repair
predictions to those which are both likely and meaningful.

One advantage to the design of this module is that it
draws upon information sources which were already part of the
system
before the introduction of the repair module.  Most of the
additional
information which the module needs was trained automatically with
statistical techniques.  The advantage to such a design is that
the module
can be easily ported to different domains with minimal additional
effort.
Another strength is that the statistical model the repair module
makes use of continually adapts during use.  This is desirable
in a
statistical approach in order to overcome problems with
unbalanced training
sets or training sets which are too small leading to over-fitting.

\section{Motivation}

The overwhelming majority of research in symbolic
approaches to handling ill-formed input has focused on flexible
parsing
strategies.   Jerry Hobbs \cite{hobbs},
David McDonald \cite{mcdonald}, Jaime Carbonell \cite{carbonell},
Wayne Ward \cite{janus},
Jill Lehman \cite{lehman}, and Alon Lavie \cite{lavie} have all
developed types of flexible
parsers.
Hobbs and McDonald each employ grammar-specific heuristics which
are suboptimal since they fall short of being completely general.
Ward and Carbonell take a pattern matching approach which is
not specific to any particular grammar but the structure of the
output
representation is not optimal for an application where the output
representation is distinct from the structure of the parse, e.\ g.\
a feature-structure, as in an interlingua-based machine translation system.

Both Lehman and Lavie take an approach which is independent of
any particular grammar and makes it possible to generate an output
representation which is distinct from the
structure of the parse.  Lehman's least-deviant-first parser can
accommodate
a wide range of repairs of parser failures.  But as it adds new
rules to its
grammar in order to accommodate idiosyncratic language patterns
it quickly
becomes intractable for multiple users.  Also, because it does
not make use
of any statistical regularities, it has to rely on heuristics to
determine
which repair to try first.  Lavie's approach is a variation on
Tomita's
Generalized LR parser which can identify and parse the maximal
subset of the
utterance which is grammatical according to its parsing grammar.
He uses
a statistical model to rank parses in order to deal with the
extraordinary
amount of ambiguity associated with flexible parsing algorithms.
His
solution is a general one.
The weakness of this approach is that part of the original
meaning of the
utterance may be thrown away with the portions of the utterance
which
were skipped in order to find a subset which can parse.

{}From a different angle, Gorin has demonstrated that it is possible
to successfully build
speech applications with a purely statistical approach.  He makes
use of
statistical correlations between features in the input and the
output which
purely symbolic approaches do not in general make use of. The
evidence
provided by each feature combines in order to calculate the output
which has
the most cumulative evidence.  In Gorin's approach, the goal is
not to
derive any sort of structural representation of the input
utterance.  It is
merely to map the set of words in the input utterance onto some
system
action.  If the goal is to map the input onto a meaning
representation, as is
the case in an interlingua-based machine translation project, the
task is
more complex.  The set of possible meaning representations even
in a
relatively small domain such as scheduling is so large that such
an approach
does not seem practical in its pure form.  But if the input
features
encode structural and semantic information, the same idea can be
used to
generate repair hypotheses.

The repair module described in this paper builds upon Lavie's and
Gorin's approaches,
reconstructing the meaning of the original utterance by
combining the fragments returned from the parser, and making
use
of statistical regularities in order to naturally determine which
combination
to try first.
In our approach
we have attempted to abstract away from any particular grammar in
order to
develop a module which could be easily ported to other domains
and other
languages.  Our approach allows the system to recover from parser
failures
and adapt without adding any extra rules to the grammar, allowing
it to
accommodate multiple users without becoming intractable.

Given a maximum of 10 questions to ask the user, it can raise the accuracy
of the parser (point value derived from automatically comparing generated
feature-structures to hand-coded ones) from 52\% to 64\% on speech data and
from 68\% to 78\% on transcribed data.  Given a maximum of 25 questions, it
can raise the accuracy to 72\% on speech-data and 86\% on transcribed data.

\section{Symbolic Information}

The system which this repair module was designed for is an
interlingua-based machine-translation system.  This means that the
goal of the analysis stage is to map the input utterance onto
a language-independent representation of meaning called an
interlingua.
Currently, the parsing grammar which is used is a semantic
grammar which
maps the input utterance directly onto the interlingua
representation.
Although the goal of an interlingua is to be language
independent, most interlinguas are domain dependent.  Although
this may seem
like a disadvantage, it actually makes it possible for domain
knowledge to be
used to constrain the set of meaningful interlingua structures for
that domain
which is particularly useful for constraining the set of possible
repairs which
can be hypothesized.  The domain which the current system focuses
on is
the scheduling domain where two speakers attempt to set up a
meeting over
the phone.

The interlingua is a hierarchical feature-structure representation.
Each level of an interlingua structure contains a frame name
which indicates
which concept is represented at that level, such as *busy or
*free.
Each frame is associated with a
set of slots which can be filled either by an atomic value or by
another
feature-structure.  At the top level, additional slots are added
for the
sentence-type and the speech-act.  Sentence-type roughly
corresponds to mood,
i.e. *state is assigned to declarative sentences and *query-if is
assigned to
yes/no questions.  The speech-act indicates what function the
utterance
performs in the discourse context.  See sample interlingua
structure in Figure~\ref{ilt}.

\begin{small}
\begin{figure}[hbt]
\begin{center}
\begin{tabbing}
(\=(speech-act \=(\=*multiple* \\
\> \> \>*state-constraint *reject)) \\
\>(sentence-type *state) \\
\>(frame *busy) \\
\>(who ((frame *i))) \\
\>(wh\=en \\
\>\>(\=(frame *special-time) \\
\>\>\>(next week)  \\
\>\>\>(specifier (*multiple* \=all-range \\
\>\>\>                       \>next))))) \\
\end{tabbing}
\end{center}
\caption{\bf Sample interlingua representation returned by the
parser for ``I'm busy all next week.''
\label{ilt}}
\end{figure}
\end{small}

The interlingua specification determines the set of possible
interlingua structures.  This specification is one of the key
symbolic
knowledge sources used for generating repair hypotheses.  It is
composed
of BNF-like rules which specify subsumption relationships between
types of
feature-structures, or between types of
feature-structures
and feature-structure specifications.

\begin{small}
\begin{figure}[hbt]
\begin{center}
\begin{tabbing}
($<TEMPORAL>$ = \=$<SIMPLE-TIME>$ \\
             \>$<INTERVAL>$ \\
             \>$<SPECIAL-TIME>$ \\
             \>$<RELATIVE-TIME>$ \\
             \>$<EVENT-TIME>$ \\
             \>$<TIME-LIST>$) \\
\end{tabbing}
\end{center}
\caption{\bf Sample interlingua specification rule for expressing a subsumption
relationship between type $<TEMPORAL>$ and more specific temporal
types.}
\end{figure}
\end{small}

\begin{small}
\begin{figure}[htb]

\begin{center}
\begin{tabbing}

($<BUSY>$ = (\=(frame *busy) \\
           \>(topic $<FRAME>$) \\
           \>(who $<FRAME>$) \\
           \>(why $<FRAME>$) \\
           \>(when $<TEMPORAL>$) \\
           \>(how-long $<LENGTH>$) \\
           \>(degree [DEGREE]))) \\

\end{tabbing}
\end{center}

\caption{\bf
Sample interlingua specification rule for expressing a subsumption
relationship between the type $<BUSY>$ and the feature-structure
specification for the frame *busy.}
\end{figure}
\end{small}

A feature-structure specification is a feature-structure who's
slots are
filled in with types rather than with atomic values or
feature-structures.
Feature-structure specifications are the leaves of the subsumption
hierarchy of interlingua specification types.

\section{Statistical Knowledge}

Intuitively, repair hypotheses are generated by computing the
mutual information between semantic grammar non-terminal symbols
and types in the interlingua specification and also between
slot/type pairs and types which are likely to be fillers of that slot.
Mutual
information is roughly a measure of how strongly associated two
concepts
are.  It is defined by the following formula:

\begin{displaymath}
log[P(c_{k} | v_{m}) / P(c_{k})]
\end{displaymath}
where $c_{k}$ is the kth element of the input vector
and $v_{m}$ is the mth element of the output vector.

Based on Gorin's approach, statistical knowledge in our repair
module is stored in a set of networks with weights which correspond to the
mutual information between an input unit and an output unit.  Gorin's
network formalism is appealing because it can be trained both off-line
with examples and on-line during use.  Another positive aspect of Gorin's
mutual information network architecture is that rather than provide a
single hypothesis about the correct output, it provides a ranked set of
hypotheses so if the user indicates that it made the wrong decision, it has
a natural way of determining what to try next.  It is also possible to
introduce new input units at any point in the training process.  This
allows the system to learn new words during use.  They will be skipped by the
parser, but the repair module can treat them like parser non-terminal symbols
and learn how to map them onto interlingua representations.  This gives the
system the
additional ability to handle nil parses.  It treats each word in the input
utterance as a chunk and proceeds as usual.  (A chunk is the Repair Module's
internal representation of a skipped portion of the input utterance.)

Our implementation of the repair module has code for generating and training
five instantiations of Gorin's network architecture, each used in a different
way in the repair process.

The first network is used for generating a set of hypothesized
types for chunks with feature-structures that have no type in the interlingua
specification.  The parse associated with these chunks is most commonly
a single symbol dominating a single word.  This symbol is used to compute
a ranked set of likely types this symbol is likely to map onto based on
how much mutual information it has with each one.  In the case that this is
a new symbol which the net has no information about yet, it will return
a ranked list of types based on how frequently those types are the correct
output.  This effect falls naturally out of the mutual information equation.

The second network is used for calculating what types are likely fillers
for particular frame slot pairs, e.\ g.\ a slot associated with a particular
frame.  This is used for generating predictions about likely
types of fillers which could be inserted in the current interlingua
structure.  This information can help the repair module interpret chunks with
uncertain types in a top-down fashion.

The third network is similar to the first network except that it maps
collections of parser non-terminal symbols onto types in the interlingua
specification.  It is used for guessing likely top level semantic frames
for sentences and for building larger chunks out of collections of
smaller ones.

The fourth network is similar to the third except instead of mapping
collections of parser non-terminal symbols onto types in the interlingua
specification, it maps them onto sentence types (see discussion on interlingua
representation).  This is used for guessing the sentence type after a new
top level semantic frame has been selected.

The fifth and final network maps a boolean value onto a ranked set of
frame slot pairs.  This is used for generating a ranked list of slots which
are likely to be filled.  This network complements the second network.
A combination of these two networks yields a list of slots which are likely
to be filled along with the types they are likely to be filled with.

My implementation of the mutual information networks allows for a mask
to filter out irrelevant hypotheses so that only the outputs which are
potentially relevant at a give time will be returned.

\section{The Repair Process: Detailed Description}

In this section I give a detailed high-level description of
the operation of the Repair Module.

\subsection{System Architecture}

The heart of the Repair Module, see Figure~\ref{sys}, is the Hypothesis
Generation Module whose purpose it is to generate repair hypotheses
which are instructions for reconstructing the speaker's meaning by
performing operations on the Chunk Structure of the parse.
The Chunk Structure represents the relationships
between the partial analysis and the analysis for each skipped
segment of the utterance.  See Figure~\ref{part}.

\begin{figure}[htb]

{\bf Speaker's Utterance:} {\em Tuesday afternoon the ninth would be okay for
me though.} \\ \\

{\bf Speech Hypothesis From the Recognizer:} Tuesday afternoon the ninth be
okay for me that. \\ \\

{\bf Partial Ananlysis:} \\

\begin{center}
\begin{tabbing}
(\=(sentence-type *fragment) \\
 \>(when (\=(frame *simple-time) \\
        \>\>(time-of-day afternoon) \\
        \>\>(day-of-week Tuesday) \\
        \>\>(day 9))) \\ \\
\end{tabbing}
\end{center}

{\bf Paraphrase of partial analysis:} Tuesday afternoon the ninth \\ \\

{\bf Skipped Portions:} \\

\begin{enumerate}

\item{((value be))}
\item{((frame *free) (who ((frame *i))) (good-bad +))}
\item{((frame *that))}

\end{enumerate}

\caption{\bf Sample Partial Parse
\label{part}}
\end{figure}

The Initialization module builds this structure
from the fragmented analysis returned by the parser.  It inserts this
structure into the Dynamic Repair Memory structure which serves as a
blackboard for communication between modules.  The Dynamic Repair Memory also
contains slots for the current repair hypothesis and the status of that
hypothesis, i.e. test, pass, fail.  There are essentially four types of repair
hypotheses that the Hypothesis Generation Module can generate.
These are guessing the
top level semantic frame for the interlingua structure of the sentence,
guessing the sentence type, combining chunks into larger chunks, and
inserting chunks into the current interlingua structure.

The Hypothesis Generation Module has access to eight different strategies for
generating repair hypotheses.  The strategy determines which of the four
types of hypotheses it should generate on each iteration.  A meta-strategy
selects which strategy to employ in a given case.

\begin{figure}[htb]
\normalpsfigureScaled{\figplain}{550pt}{575pt}{.45}{.45}{sys2.eps}
\caption{\bf Repair Module System Architechture
\label{sys}}
\end{figure}

Once the hypothesis
is generated, it is sent to the Question Generation Module which generates a
question for the user to check whether the hypothesis is correct.  After the
user responds, the status of the hypothesis is noted in the Dynamic Repair
Memory and if the response was positive, the Interlingua Update Module makes
the specified repair and updates the Dynamic Repair Memory structure.
It is the Interlingua Update Module which uses these hypotheses
to actually make the repairs in order to derive the
complete meaning representation for the utterance from the partial
analysis and the analysis for the skipped portions.

If the status indicates that the speaker's response was negative,
the Hypothesis Generation Module will suggest an alternative
repair hypothesis which is possible since the mutual information nets
return a ranked list of predictions rather than a single one.  In this way the
repair module negotiates with the speaker about what was meant until an
acceptable interpretation can be constructed.  See Figure~\ref{done}.
When the goal returns
positive, the networks are reinforced with the new information so they can
improve their performance over time.

\begin{figure}[hbt]

\begin{center}
\begin{tabbing}

{\bf Interlingua Representation:} \\ \\

(\=(sentence-type *state) \\
 \>(frame *free) \\
 \>(who ((frame *i))) \\
 \>(when (\=(frame *simple-time) \\
        \>\>(time-of-day afternoon) \\
        \>\>(day-of-week Tuesday) \\
        \>\>(day 9)))) \\ \\

{\bf Paraphrase:} I am free Tuesday afternoon the ninth.

\end{tabbing}
\end{center}

\caption{\bf Complete Meaning Representation After Repair
\label{done}}
\end{figure}

\subsection{The Three Questions}

The eight strategies are generated by all possible ways of selecting either
top-down or bottom-up as the answer to three questions.

The first question is, ``What will be the top level semantic frame?''.  The
top-down approach is to keep the partial analysis returned by the parser as
the top level structure thereby accepting the top level frame in the
partial analysis returned by the parser as representing the gist of the
meaning of the sentence.  The bottom-up approach is to assume that the
partial analysis returned by the parser is merely a portion of the meaning
of the sentence which should fit into a slot inside of some other top level
semantic frame.  This is the case in the example in Figure~\ref{part}.

If bottom-up is selected, a new top level semantic frame is chosen
by taking the set of all parser non-terminal symbols
in the tree structure for the partial analysis and from each skipped
segment and computing the mutual information between that set and each
interlingua specification type.  This gives it a ranked set of possible types
for the top level interlingua structure.  The interlingua specification
rule for the selected type would then become the template for fitting in the
information extracted from the partial analysis as well as from the skipped
portions of the utterance.  See Figure~\ref{1st}.
If a new top-level frame was guessed, then a new
sentence-type must also be guessed.  Similar to guessing
a top level frame, it computes the mutual information between the
same set of parser non-terminal symbols and the set of sentence-types.

\begin{figure}[hbt]

{\bf Question:} What will be the top level structure? \\

{\bf Answer:} Try Bottom-Up. \\  \\

{\bf Hypothesis:} (top-level-frame ((frame-name *free))) \\

{\bf Question:} {Is your sentence mainly about someone being free?} \\

{\bf User Response:} {\em Yes.} \\ \\

{\bf New Current Interlingua Structure:} \\

((frame *free)) \\ \\

{\bf Skipped Portions:} \\

\begin{enumerate}

\item{((value be))}
\item{((frame *free) (who ((frame *i))) (good-bad +))}
\item{((frame *that))}
\item{((frame *simple-time) (time-of-day afternoon) (day-of-week Tuesday) (day
9))}

\end{enumerate}

\caption{\bf The First Question
\label{1st}}
\end{figure}

The second question is, ``How will constituents be built?''.  The
top-down approach is to assume that a meaningful constituent to insert into
the current interlingua structure for the sentence can be found by simply
looking at available chunks and portions of those chunks.  See
Figure~\ref{2nd}.  The bottom-up
approach is to assume that a meaningful chunk can be constructed by
combining chunks into larger chunks which incorporate their meaning.
The process of generating predictions about how to combine chunks into
larger chunks is similar to guessing a top-level frame from the utterance
except that only the parser non-terminal symbols for the segments in question
are used to make the computation.

\begin{figure}[hbt]

{\bf Question:} How will constituents be built? \\

{\bf Answer:} Try Top-Down. \\  \\

{\bf Available Chunks:} \\

\begin{enumerate}

\item{((value be))}
\item{((frame *free) (who ((frame *i))) (good-bad +))}
\item{((frame *that))}
\item{((frame *simple-time) (time-of-day afternoon) (day-of-week Tuesday) (day
9)) \\}

\end{enumerate}

{\bf Constituents:} \\

\begin{enumerate}

\item{((frame *simple-time) (time-of-day afternoon) (day-of-week Tuesday) (day
9))}

\item{((frame *free) (who ((frame *i))) (good-bad +))}

\item{((frame *i))}

\item{((frame *that))}

\item{((value be))}

\end{enumerate}

\caption{\bf The Second Question
\label{2nd}}
\end{figure}

The third question is, ``What will drive the search process?''.  The
bottom-up approach is to generate predictions of where to insert chunks
by looking at the chunks themselves and determining where in the interlingua
structure they might fit in.  See Figure~\ref{3rd1}.

\begin{figure}[hbt]

{\bf Question:} What will drive the search process? \\

{\bf Answer:} Try Bottom-Up. \\

{\bf Current Constituent:}

\begin{center}
\begin{tabbing}
 (\=(frame *simple-time) \\
  \>(time-of-day afternoon) \\
  \>(day-of-week Tuesday) \\
  \>(day 9))) \\
\end{tabbing}
\end{center}

{\bf Hypothesis:} \\

\begin{center}
\begin{tabbing}
(frame-slot (\=(frame-name *free) \\
             \>(when (\=(frame *simple-time) \\
             \>       \>(time-of-day afternoon) \\
             \>       \>(day-of-week Tuesday) \\
             \>       \>(day 9))))) \\
\end{tabbing}
\end{center}

{\bf Question:} {Is Tuesday afternoon the ninth the time of being free in
your sentence?} \\

{\bf User Response:} {\em Yes.} \\ \\

{\bf New Current Interlingua Structure:} \\

\begin{center}
\begin{tabbing}
(\=(sentence-type *state) \\
 \>(frame *free) \\
 \>(when (\=(frame *simple-time) \\
        \>\>(time-of-day afternoon) \\
        \>\>(day-of-week Tuesday) \\
        \>\>(day 0)))) \\ \\
\end{tabbing}
\end{center}
\caption{\bf The Third Question - Part 1
\label{3rd1}}
\end{figure}

The top-down approach is to look at the
interlingua structure, determine what slot is likely to be filled in, and look
for a chunk which might fill that slot.  See Figure~\ref{3rd2}.

\begin{figure}[hbt]

{\bf Question:} What will drive the search process? \\

{\bf Answer:} Try Top-Down. \\

{\bf Current Slot:} who \\

{\bf Hypothesis:} (frame-slot ((frame-name *free) (who ((frame *i))))) \\

{\bf Question:} Is it ``I'' who is being free in your sentence? \\

{\bf User Response:} {\em Yes.}

{\bf New Current Interlingua Structure:} \\

\begin{center}
\begin{tabbing}
(\=(sentence-type *state) \\
 \>(frame *free) \\
 \>(who ((frame *i))) \\
 \>(when (\=(frame *simple-time) \\
        \>\>(time-of-day afternoon) \\
        \>\>(day-of-week Tuesday) \\
        \>\>(day 0)))) \\ \\
\end{tabbing}
\end{center}

\caption{\bf The Third Question - Part 2
\label{3rd2}}
\end{figure}

The difference between these
strategies is primarily in the ordering of hypotheses.  But there is also
some difference in the breadth of the search space.  The bottom-up
approach will only generate hypotheses about chunks which it has.  And if
there is some doubt about what the type of a chunk is, only a finite number
of possibilities will be tested, and none of these may match something which
can be inserted into one of the available slots.  The top-down approach
generates its predictions based on what is likely to fit into available
slots in the current interlingua structure.  It first tries to find a likely
filler which matches a chunk which
has a definite type, but in the absence of this eventuality, it will assume
that a chunk with no specific type is whatever type it guesses can fit into
a slot.  And if the user confirms that this slot should be filled with this
type, it will learn the mapping between the symbols in that chunk and that
type.  Learning new words is more likely to occur with the top-down approach
than with the bottom-up approach.

The meta-strategy answers these questions, selecting the strategy to
employ at a given time.  Once a strategy is selected, it continues until
it either makes a repair or cannot generate anymore questions given the
current state of the Dynamic Repair Memory.  Also, once the first question
is answered, it is never asked again since once the top level frame is
confirmed, it can be depended upon to be correct.

The meta-strategy attempts to answer the first question at the beginning of
the search process.  If the whole input utterance parses or the parse
quality indicated by the parser is good and the top level frame guessed
as most likely by the mutual information nets matches the one chosen by
the parser, it assumes it should take the top-down approach.  If the parse
quality is bad, it assumes it should guess a new top level frame, but it
does not remove the current top level frame from its list of possible
top level frames.  In all other cases, it confirms with the user whether
the top level frame selected by the parser is the correct one and if it
is not, then it proceeds through its list of hypotheses until it locates
the correct top level frame.

Currently, the meta heuristic always answers the second question the
same way.  Preliminary results indicated that in the great majority of
cases, the repair module was more effective when it took the top down
approach.  It is most often the case that the chunks which are needed
can be located within the structures of the chunks returned by the parser
without combining them.  And even when it is the case that chunks should
be combined in order to form a chunk which fits into the current interlingua
structure, the same effect can be generated by mapping the top level
structure of the would be combined chunk onto an available chunk with
an uncertain type and then inserting the would be constituent chunks into
this hypothesized chunk later.  Preliminary tests indicated that the option
of combining chunks only yielded an increase in accuracy in about 1\% of the
129 cases tested.  Nevertheless, it would be ideal for the meta heuristic to
sense when it is likely to be useful to take this approach, no matter how
infrequent.  This will be a direction for future research.

The third question is answered by taking the bottom-up approach early,
considering only chunks with a definite type and then using a top down
approach for the duration of the repair process for the current interlingua
structure.

The final task of the meta heuristic is for it to decide when to stop
asking questions.  Currently it does this when there are no open slots or
it has asked some arbitrary maximum number of questions.  An important
direction for future research is to find a better way of doing this.
Currently, the repair module asks primarily useful questions (yielding
an increase in accuracy) early (within the first 5 or 10 questions) and
then proceeds to ask a lot of irrelevant questions.  But I have not found
an optimal maximum number of questions.  If the number of questions is
too small, it will not be able to learn some new input patterns and
sometimes fails to recover information it would have been able to recover
had it been allowed to ask a few more questions.  But if the
number is too large, it is unnecessarily annoying for the user, particularly in
cases where the important information was recovered early in the process.

\subsection{User Interaction}

User interaction is an essential part of our approach.  The ideal in
speech-to-speech translation has been direct through-put from input
speech to output speech.  But this leaves the speaker with no idea
of what the system understood from what was said or what is
ultimately communicated to the other speaker.  This is particularly
a problem with flexible parsing techniques where the parser must take
some liberties in finding a parse for ill-formed input.

Because our Hypothesis Generation Module makes hypotheses about local
repairs, the questions generated focus on local information in the
meaning representation of the sentence.  For instance, rather than confirm
global meaning represenations as in , ``Did you mean to say X?'', it
confirms local information as in, ``Is two o'clock the time of being busy
in your sentence?'' which confirms that the representation for ``two o'clock''
should be inserted into the {\em when} slot in the *busy frame.

\section{Results}

\begin{figure}[hbt]
\normalpsfigureScaled{\figplain}{600pt}{800pt}{.45}{.4}{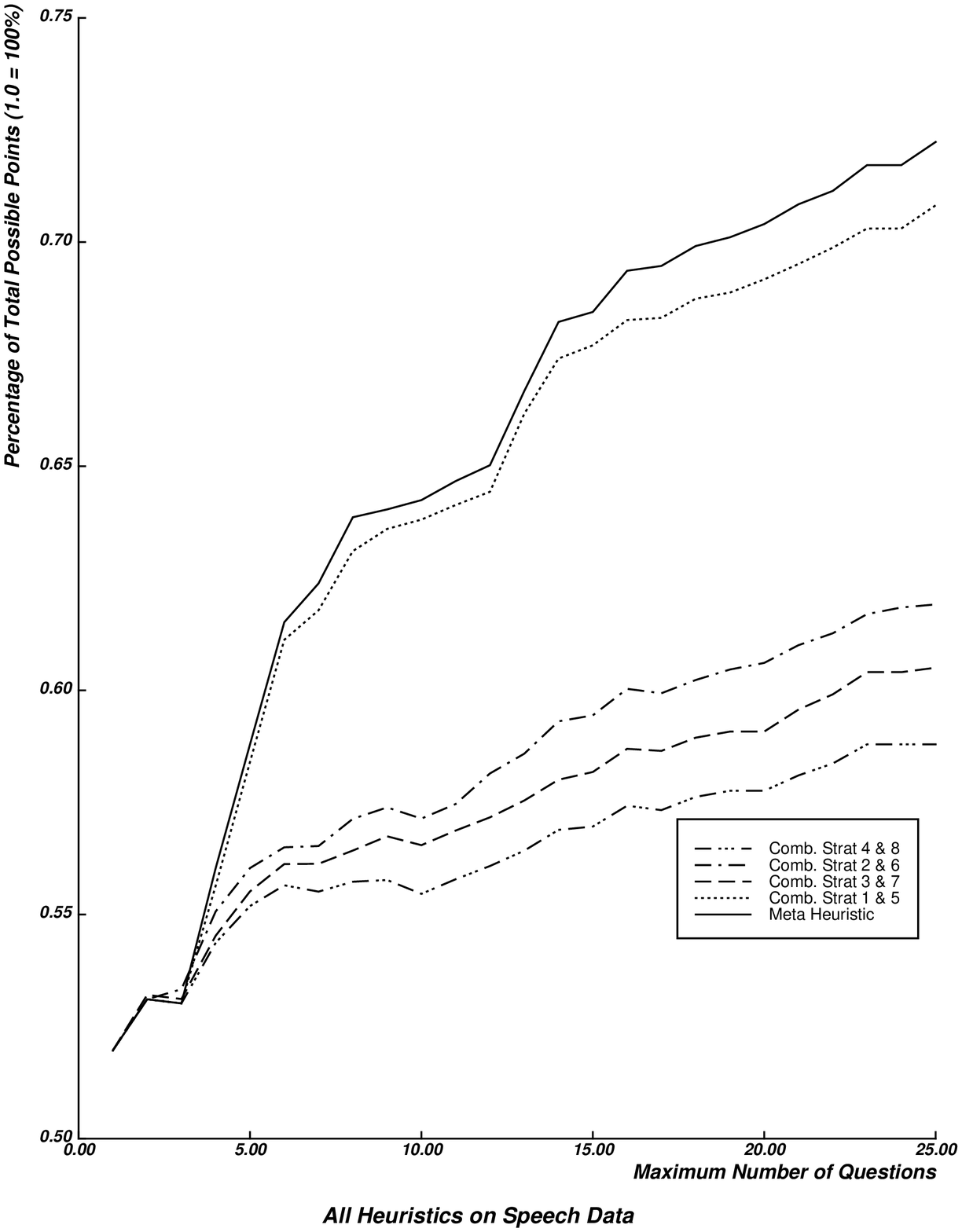}
\caption{\bf Results from All Strategies on Speech Data
\label{speech}}
\end{figure}

Figure~\ref{speech} displays the relative performance of the eight strategies
compared to the meta strategy on speech data.

Given a maximum of 10 questions to ask the user,
the repair module can raise the accuracy
of the parser (point value derived from automatically comparing generated
feature-structures to hand-coded ones) from 52\% to 64\% on speech data and
from 68\% to 78\% on transcribed data.  Given a maximum of 25 questions, it
can raise the accuracy to 72\% on speech-data and 86\% on transcribed data.

\section{Conclusions and Future Directions}

This document describes an approach to interactive repair of fragmented
parses in the context of a speech-to-speech translation project of significant
scale. It makes it possible to use symbolic knowledge sources to the extent
that they are available and uses statistical knowledge to fill in the gaps.
This gives it the ability to keep the preciseness of symbolic approaches
wherever possible as well as the robustness of statistical approaches
wherever symbolic knowledge sources are not available.  It is a
general approach which applies regardless of how degraded the input is, even
if the sentence completely fails to parse.

The primary weakness of this approach is that it relies too heavily on
user interaction.  One goal for future research will be to look
into various ways of reducing this burden on the user.  The following is a list
of potential avenues of exploration:

\begin{enumerate}

\item{Reduce unnecessary positive confirmations by developing a reliable
confidence measure.}

\item{Use contextual knowledge and possibly some domain knowledge to eliminate
hypotheses which don't make sense.}

\item{Develop heuristics for rejecting sentences which are out of domain.}

\item{Introduce a mechanism for enforcing global constraints, i.\ e.\
agreement, and other selectional restrictions.}

\end{enumerate}

\end{document}